\documentclass[10pt,twocolumn,letterpaper]{article}
\pdfoutput=1
\usepackage{times}
\usepackage{epsfig}
\usepackage{graphicx}
\usepackage{subcaption}
\usepackage{amsmath}
\usepackage{amssymb}
\usepackage{tikz}
\usepackage{multirow}
\usetikzlibrary{decorations.pathreplacing,fit,calc}
\usepackage{multirow,tabularx}
\usepackage[bookmarks=false]{hyperref}

\begin{document}

\title{eXclusive Autoencoder (XAE) for Nucleus Detection and Classification on Hematoxylin and Eosin (H\&E) Stained Histopathological Images}

\author{Chao-Hui HUANG\\
Pfizer Inc. \\
{\tt\small \url{http://www.chhuang.org}}
\and
Daniel RACOCEANU\\
Sorbonne Universit\'{e}, Paris, France\\
{\tt\small daniel.racoceanu@sorbonne-universite.fr}
}

\maketitle

\begin{abstract}

In this paper, we introduced a novel feature extraction approach, named exclusive autoencoder (XAE), which is a supervised version of autoencoder (AE), able to largely improve the performance of nucleus detection and classification on hematoxylin and eosin (H\&E) histopathological images. The proposed XAE can be used in any AE-based algorithm, as long as the data labels are also provided in the feature extraction phase. In the experiments, we evaluated the performance of an approach which is the combination of an XAE and a fully connected neural network (FCN) and compared with some AE-based methods. For a nucleus detection problem (considered as a nucleus/non-nucleus classification problem) on breast cancer H\&E images, the F-score of the proposed XAE+FCN approach achieved $96.64\%$ while the state-of-the-art was at $84.49\%$. For nucleus classification on colorectal cancer H\&E images, with the annotations of four categories of epithelial, inflammatory, fibroblast and miscellaneous nuclei. The F-score of the proposed method reached $70.4\%$. We also proposed a lymphocyte segmentation method. In the step of lymphocyte detection, we have compared with cutting-edge technology and gained improved performance from $90\%$ to $98.67\%$. We also proposed an algorithm for lymphocyte segmentation based on nucleus detection and classification. The obtained Dice coefficient achieved $88.31\%$ while the cutting-edge approach was at $74\%$.


\end{abstract}


\section{Introduction}
\label{sec:introduction}

Nucleus detection and classification are critical steps for computer-aided pathology \cite{Rujuta:17}. \textit{E.g.}, in the Nottingham breast cancer grading system, 2 of 3 criteria are nuclei-related \cite{Balslev:94:281}. 
Although immuno-histo-chemical (IHC) staining may produce images which are easier for nucleus detection. However, the conventional hematoxylin \& eosin (H\&E) staining remains the golden standard for cancer diagnosis \cite{Balslev:94:281}.

Automated nucleus detection, essentially, is a nucleus/non-nucleus classification problem, which is not a straightforward task due to: 1) the complexity and variation of their appearances under a microscope; and 2) a large number of nuclei in a whole slide image. 
Recently, there has been interesting in deep learning and their applications on biomedical image analysis, including nucleus detection. \textit{E.g.}, Xu \textit{et al.} \cite{Xu:16:119} proposed methods of stacked sparse autoencoder (SSAE); Khoshdeli \textit{et al.} \cite{Khoshdeli:17:105} reported convolutional neural network (CNN) based approaches.

Nucleus classification is a further challenge which includes issues: 1) the similarity between the images of various nucleus categories; 2) a large number of nuclei in a whole slide image. In order to tackle these problems, a more complicate (thus, more computational costly) approach is usually need. \textit{E.g.}, Sirinukunwattana \textit{et al.} proposed a locality sensitive deep learning approach for nucleus detection and classification \cite{Sirinukunwattana:16:1196}, which took the advantages of location sensitivity on the given images. 

Many studies of automated nucleus detection and classification involve a two-step approach, which is, first, a feature extraction step, following by a classification step. An conventional autoencoder (AE) is frequently being used at the step of feature extraction. 

A conventional AE is a 3-layer NN which learns to represent the input values in the hidden units with a relatively lower dimensional data space. Then, by transferring the values of the hidden units to the output with the same data dimension of the input, the values can be reconstructed. 

Many AE-based nucleus detection/classification methods have achieved excellent results (\textit{e.g.}, \cite{Xu:16:119, Sirinukunwattana:16:1196}). However, as a unsupervised neural network (NN), a conventional AE has its limitations, including: 1) AE does not provide clear clues indicating that a obtained feature is belonging to which class; 2) a feature obtained using a conventional AE may be appeared in two or more classes of the given dataset, thus, the feature will not be able to contribute to the  steps of classification; and 3) in a unbalanced dataset, the feature space may be occupied by the majority of the dataset, as a result, the subtle (but critical) features may be ignored.

Thus, we propose an improved version of AE, called exclusive autoencoder (XAE), which aims to learn not only the mutual features of some (or all) classes, but also the exclusive features of each class of the given dataset based on the corresponding label set. The idea is, given a class of the dataset, the following classification task is related to a joint probability of all hidden unit activation. By identifying the exclusive components of each class of the given dataset, it is possible to minimize the divergence of two joint probabilities of the hidden units activated by two different data classes. As a result, the performance of the following classification task can be improved.

In this paper, we will discuss how does the proposed XAE largely improved the existing AE-based nucleus detection/classification. In the following sections, we will, first, introduce the approach of XAE (Sect.~\ref{sec:method}), including a toy example classification using MNIST \cite{MNISTHandwrittenDigitDatabase}. Then, in the result section, the examples of using XAE 
on nucleus detection, classification, and a real world application of lymphocyte segmentation based on the proposed approaches will be presented (Sect.~\ref{sec:results}). Finally, the conclusions will be drawn (Sect.~\ref{sec:conclusions}).

\section{Method}
\label{sec:method}

\begin{figure}
	\begin{center}
		\begin{subfigure}[b]{0.5\textwidth}	
			\centering
			\begin{tikzpicture}[shorten >=1pt,->,draw=black!50, node distance=0cm,thick,scale=0.7]
			\tikzstyle{every pin edge}=[<-,shorten <=1pt]
			\tikzstyle{neuron}=[circle,fill=black!25,minimum size=17pt,inner sep=0pt]
			\tikzstyle{group}=[rectangle,draw=black!50,rounded corners=4pt,inner sep=2pt]
			\tikzstyle{input neuron}=[neuron, fill=magenta!100, style={scale=0.6}]
			\tikzstyle{output neuron}=[neuron, fill=magenta!100, style={scale=0.6}]
			\tikzstyle{hidden neuron}=[neuron, fill=red!100, style={scale=0.6}]
			\tikzstyle{bias neuron}=[neuron, fill=black!100, style={scale=0.6}]
			\tikzstyle{annot} = [text width=4em, text centered, style={scale=0.6}]
			
			\foreach \name / \y in {1,...,4}
			\node[input neuron] (I-\name) at (0,-\y) {};
			\node[group, pin=left:\scalebox{0.8}{$\mathbf{x}_{k}$}, fit=(I-1)(I-2)(I-3)(I-4)] (GI-1) {};
			\node[bias neuron, yshift=0cm, label={[label distance=-0.2cm]270:\scalebox{0.8}{$\mathbf{b}_{\text{encoder}}$}}] (B-1) at (1cm,-5) {};
			
			\foreach \name / \y in {1,...,2}
			\node[hidden neuron, yshift=-1cm] (H-\name) at (3cm,-\y) {};
			\node[group, fit=(H-1)(H-2), label={[label distance=-0.1cm]87:\scalebox{0.8}{$\mathbf{z}_{k}$}}] (GH-1) {};
			\node[bias neuron, yshift=0cm, label={[label distance=-0.2cm]270:\scalebox{0.8}{$\mathbf{b}_{\text{decoder}}$}}] (B-2) at (4cm,-5) {};
			
			\foreach \name / \y in {1,...,4}
			\node[output neuron] (O-\name) at (6cm,-\y) {};
			\node[group, pin=right:\scalebox{0.8}{$\mathbf{y}_{k}$}, fit=(O-1)(O-2)(O-3)(O-4)] (GO-1) {};
			
			\path (GI-1) edge (GH-1);
			\path (B-1) edge (GH-1);
			\path (GH-1) edge (GO-1);
			\path (B-2) edge (GO-1);
			\end{tikzpicture}
			\caption{A conventional AE.}
			\label{fig:ae}
		\end{subfigure}
		\begin{subfigure}[b]{0.5\textwidth}
			\centering
			\begin{tikzpicture}[shorten >=1pt,->,draw=black!50, node distance=0cm,thick,scale=0.7]
			\tikzstyle{every pin edge}=[<-,shorten <=1pt]
			\tikzstyle{neuron}=[circle,fill=black!25, line width=1mm, minimum size=17pt,inner sep=0pt]
			\tikzstyle{group}=[rectangle,draw=black!100,rounded corners=4pt,inner sep=2pt]
			\tikzstyle{input neuron class 1}=[neuron, fill=magenta!100, style={scale=0.6}]
			\tikzstyle{input neuron class 2}=[neuron, fill=cyan!100, style={scale=0.6}]
			\tikzstyle{output neuron class 1}=[neuron, fill=magenta!100, style={scale=0.6}]
			\tikzstyle{output neuron class 2}=[neuron, fill=cyan!100, style={scale=0.6}]
			\tikzstyle{hidden neuron exclusive 1}=[neuron, fill=red!100, style={scale=0.6}]
			\tikzstyle{hidden neuron exclusive 2}=[neuron, fill=green!100, style={scale=0.6}]
			\tikzstyle{hidden neuron mutual}=[neuron, fill=blue!100, style={scale=0.6}]
			\tikzstyle{bias neuron}=[neuron, fill=black!100, style={scale=0.6}]
			\tikzstyle{annot} = [text width=4em, text centered, style={scale=0.6}]
			
			\foreach \name / \y in {1,...,4}
			\node[input neuron class 1] (I-\name) at (0,-\y) {};
			\node[group, pin=left:\scalebox{0.8}{$\mathbf{x}_{k}^{(\mathbf{S}_{0})}$}, fit=(I-1)(I-2)(I-3)(I-4)] (GI-1) {};
			
			\foreach \name / \y in {5,...,8}
			\node[input neuron class 2] (I-\name) at (0,-\y) {};
			\node[group, pin=left:\scalebox{0.8}{$\mathbf{x}_{k}^{(\mathbf{S}_{1})}$}, fit=(I-5)(I-6)(I-7)(I-8)] (GI-2) {};	
			\node[bias neuron, yshift=0cm, label={[label distance=-0.2cm]270:\scalebox{0.8}{$\mathbf{b}_{\text{encoder}}$}}] (B-1) at (1cm,-9) {};
			
			\foreach \name / \y in {1,...,2}
			\node[hidden neuron exclusive 1, yshift=0cm] (H-\name) at (3cm,-\y) {};
			\node[group, fit=(H-1)(H-2), label={[label distance=-0.1cm]87:\scalebox{0.8}{$\mathbf{z}_{k}^{(\mathbf{S}_{i},\mathbf{T}_{0})}$}}] (GH-1) {};
			
			\foreach \name / \y in {4,...,5}
			\node[hidden neuron mutual, yshift=0cm] (H-\name) at (3cm,-\y) {};
			\node[group, fit=(H-4)(H-5), label={[label distance=-0.1cm]87:\scalebox{0.8}{$\mathbf{z}_{k}^{(\mathbf{S}_{i},\mathbf{T}_{1})}$}}] (GH-2) {};
			
			\foreach \name / \y in {7,...,8}
			\node[hidden neuron exclusive 2, yshift=0cm] (H-\name) at (3cm,-\y) {};
			\node[group, fit=(H-7)(H-8), label={[label distance=-0.1cm]87:\scalebox{0.8}{$\mathbf{z}_{k}^{(\mathbf{S}_{i},\mathbf{T}_{2})}$}}] (GH-3) {};	
			\node[bias neuron, yshift=0cm, label={[label distance=-0.2cm]270:\scalebox{0.8}{$\mathbf{b}_{\text{decoder}}$}}] (B-2) at (4cm,-9) {};
			
			\foreach \name / \y in {1,...,4}
			\node[output neuron class 1] (O-\name) at (6cm,-\y) {};
			\node[group, pin=right:\scalebox{0.8}{$\mathbf{y}_{k}^{(\mathbf{S}_{0})}$}, fit=(O-1)(O-2)(O-3)(O-4)] (GO-1) {};
			
			\foreach \name / \y in {5,...,8}
			\node[output neuron class 2] (O-\name) at (6cm,-\y) {};
			\node[group, pin=right:\scalebox{0.8}{$\mathbf{y}_{k}^{(\mathbf{S}_{1})}$}, fit=(O-5)(O-6)(O-7)(O-8)] (GO-2) {};
			
			\path (GI-1) edge (GH-1);
			\path (GI-1) edge (GH-2);
			\path (GI-2) edge (GH-2);
			\path (GI-2) edge (GH-3);
			\path (B-1) edge (GH-1);
			\path (B-1) edge (GH-2);
			\path (B-1) edge (GH-3);
			\path (GH-1) edge (GO-1);
			\path (GH-2) edge (GO-1);
			\path (GH-2) edge (GO-2);
			\path (GH-3) edge (GO-2);
			\path (B-2) edge (GO-1);
			\path (B-2) edge (GO-2);
			\end{tikzpicture}
			\caption{An XAE for a 2-class dataset.}
			\label{fig:xae}
		\end{subfigure}
	\end{center}
	\caption{The comparison between a conventional AE and XAE. (a) a conventional AE and (b) an XAE aiming to identify the mutual components of the dataset, as well as the exclusive components for each class in the dataset. Note that these two diagrams do not imply the data dimension in the XAE is twice larger than the AE. Instead, (b) represents the fact that the XAE updates its weights based on the corresponding label set $\mathbf{S}_{i}$ of the input $\mathbf{x}_{k}^{(\mathbf{S}_{i})}$, and in this case, $i\in\left\lbrace 0,1\right\rbrace$.}
	\label{fig:ae_and_xae}
\end{figure}
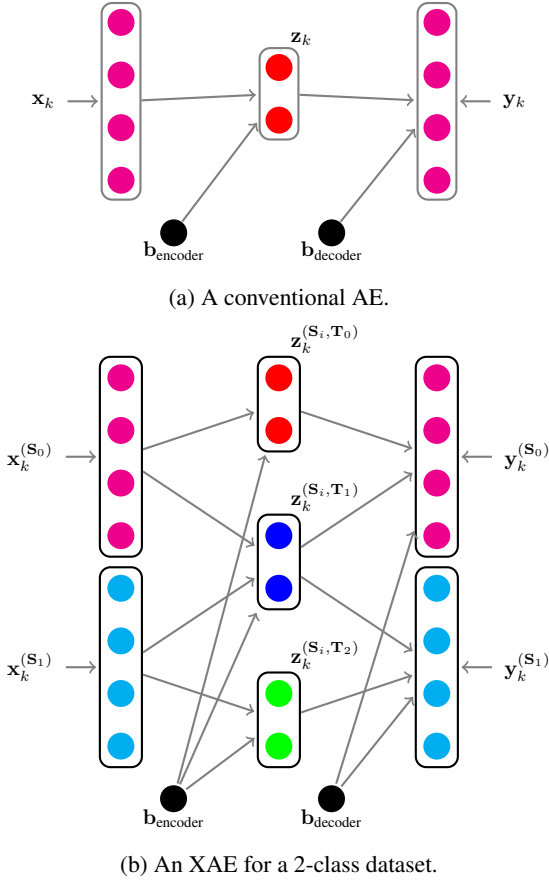

AE is a 3-layer NN aims to reconstruct the input by a smaller number of hidden units. In other words, AE is able to extract the key components of the given training patterns. This capability has been widely used as a pre-training step of feature extraction. \textit{E.g.,} sparse autoencoders \cite{Ng}, in which, the output of the hidden layer is sparse, and each unit in the hidden represents a unique feature of the given training patterns. The performance of the following classification depends on a probability that one or more hidden units are correlated to the data labels. However, there is no guarantee for this probability. In other words, if this probability can be controlled, the accuracy of the following classification can be improved. 

One idea is, given a dataset of two classes, we can partition the hidden layer of AE into two segments, so that given the training patterns of the first data class, the first segment is highly correlated to the first segment, but uncorrelated to the second segment, and \textit{vice visa}. More precisely, given the training patterns of the first data class, we wish the units in the first segment can represent all critical features, on the other hand, the units in the second segment are all $0$. However, this is a very strict requirement. Thus, we pursuit the next best option, that is, making the first segment sparse and the second segment falling into a narrow zero-mean Gaussian distribution. To achieve this goal, we introduce XAE.


The conceptual idea of XAE is to partition the hidden layer into two or more segments so that a segment is only correlated to relevant sub-dataset, according to the given labels. There are two types of hidden layer segments: 1) the segments for exclusive components (\textit{e.g.}, $\mathbf{z}_{k}^{(\mathbf{S}_{i},\mathbf{T}_{0})}$ and $\mathbf{z}_{k}^{(\mathbf{S}_{i},\mathbf{T}_{2})}$ in Fig.~\ref{fig:xae}), which represents the feature representation for a specific class of the dataset; and 2) the segments for mutual components (\textit{e.g.}, $\mathbf{z}_{k}^{(\mathbf{S}_{i},\mathbf{T}_{1})}$ in Fig.~\ref{fig:xae}), representing the mutual features across two or more data classes. The XAE configuration can be expanded for datasets with two or more data categories. In the following sections, the further details of XAE architecture will be discussed.

\subsection{Notations for Labels and Datasets:} Given a dataset with two or more data classes, the labels of the data classes can be described as a finite set,
\begin{equation}
\mathbf{L}=\left\lbrace l_{1},l_{2},\cdots,l_{i},\cdots,l_{|\mathbf{L}|}\right\rbrace,
\end{equation}
where each $l_{i}$ represents a label in the problem. The cardinality, $|\mathbf{L}|$, is the total number of label types in the problem. In some situations, a dataset may or may not contain data instances corresponding to multiple labels \cite{Zhou:12:2291}. Thus, instead of using a single label $l_{i}$ for indicating the category of a dataset, we can use a set of labels, $\mathbf{S}_{i}$, to describe the involved labels of the $i^{th}$ group of data instances, which share the same label sets, as the follows:
\begin{equation}
\mathbf{S}_{i}\subseteq\mathbf{L},\mathbf{S}_{i}\neq\emptyset.
\end{equation}
Thus, given a dataset, $\mathbf{X}=\lbrace\mathbf{x}_{1},\mathbf{x}_{2},\cdots,\mathbf{x}_{k},\cdots\rbrace$, each $\mathbf{x}_{k}\in\mathbb{R}^{M}$ is a $M$ dimensional data vector. The subset of all labels, $\mathbf{S}_{i}$, can be used to indicate the categories of a sub-dataset, $\mathbf{X}^{\scalebox{0.66}{$(\mathbf{S}_{i})$}}\subseteq\mathbf{X}$, representing in the form of a block matrix:
\begin{equation}
\mathbf{X}^{\scalebox{0.66}{$(\mathbf{S}_{i})$}}=\left[\mathbf{x}_{1}^{\scalebox{0.66}{$(\mathbf{S}_{i})$}},\mathbf{x}_{2}^{\scalebox{0.66}{$(\mathbf{S}_{i})$}},\cdots,\mathbf{x}_{k}^{\scalebox{0.66}{$(\mathbf{S}_{i})$}},\cdots\right]\in\mathbb{R}^{M\times |\mathbf{X}^{\scalebox{0.66}{$(\mathbf{S}_{i})$}}|},
\label{eq:XSi}
\end{equation}
where each $\mathbf{x}_{k}^{\scalebox{0.66}{$(\mathbf{S}_{i})$}}$ represents a data instance corresponding to one or more data categories according to the defined $\mathbf{S}_{i}$. In other words, $\mathbf{S}_{i}$ represents the \textit{priori knowledge} of $\mathbf{X}^{\scalebox{0.66}{$(\mathbf{S}_{i})$}}$.

\subsection{Architecture of XAE}
\paragraph{Encoder:} Given the input $\mathbf{x}_{k}^{\scalebox{0.66}{$(\mathbf{S}_{i})$}}$, the hidden layer is obtained by
\begin{equation}
\mathbf{z}_{k}^{\scalebox{0.66}{$(\mathbf{S}_{i})$}}=a_{\text{encoder}}\left(\mathbf{W}^{T}\mathbf{x}_{k}^{\scalebox{0.66}{$(\mathbf{S}_{i})$}}+\mathbf{b}_{\text{encoder}}\right),
\label{eq:zkSi}
\end{equation}
where $a_{\text{encoder}}(\cdot)$ represents an activation function for the encoder;
$\mathbf{b}_{\text{encoder}}\in\mathbb{R}^{N}$ is the bias term for the encoder;
$\mathbf{z}_{k}^{\scalebox{0.66}{$(\mathbf{S}_{i})$}}$ is the output of the hidden layer;
and $\mathbf{W}$ is the feature set representing in the form of a block matrix, which, in the fact, is a concatenation of some subset of all features, \textit{e.g.},  
\begin{align}
\mathbf{W}&=\left[\mathbf{W}_{(M\times n_{1})}^{\scalebox{0.66}{$(\mathbf{T}_{1})$}}~\mathbf{W}_{(M\times n_{2})}^{\scalebox{0.66}{$(\mathbf{T}_{2})$}}~\cdots\right]\in\mathbb{R}^{M\times N},\text{ and }\nonumber\\
N&=\sum_{j}n_{j}.
\label{eq:W}
\end{align}
Each $\mathbf{T}_{j}\subseteq\mathbf{L}$, which is similarly defined as of $\mathbf{S}_{i}$. However, $\mathbf{T}_{j}$ and $\mathbf{S}_{i}$ are serving different purposes. That is, each $\mathbf{S}_{i}$ is used as a meta-variable, mapping $\mathbf{X}^{\scalebox{0.66}{$(\mathbf{S}_{i})$}}$ to one or more labels, while 
each $\mathbf{T}_{j}$ represents the desired \textit{segmentation} of $\mathbf{W}^{\scalebox{0.66}{$(\mathbf{T}_{j})$}}$, correlating each $\mathbf{W}^{\scalebox{0.66}{$(\mathbf{T}_{j})$}}$ to one or more data classes.

By following this concept, given a set of $\mathbf{W}^{\scalebox{0.66}{$(\mathbf{T}_{j})$}}$, $\mathbf{z}_{k}^{\scalebox{0.66}{$(\mathbf{S}_{i})$}}$ is equivalent to a concatenation:
\begin{equation}
\mathbf{z}_{k}^{\scalebox{0.66}{$(\mathbf{S}_{i})$}}=\left[\left(\mathbf{z}_{k}^{\scalebox{0.66}{$(\mathbf{S}_{i},\mathbf{T}_{1})$}}\right)^{T}~\left(\mathbf{z}_{k}^{\scalebox{0.66}{$(\mathbf{S}_{i},\mathbf{T}_{2})$}}\right)^{T}~\cdots\right]^{T}\in\mathbb{R}^{N},
\end{equation}
where each
\begin{equation}
\mathbf{z}_{k}^{\scalebox{0.66}{$(\mathbf{S}_{i},\mathbf{T}_{j})$}}=\left[z_{k,1}^{\scalebox{0.66}{$(\mathbf{S}_{i},\mathbf{T}_{j})$}},z_{k,2}^{\scalebox{0.66}{$(\mathbf{S}_{i},\mathbf{T}_{j})$}},\cdots,z_{k,n_{j}}^{\scalebox{0.66}{$(\mathbf{S}_{i},\mathbf{T}_{j})$}}\right]^{T}\in\mathbb{R}^{n_{j}},
\label{eq:zkSiTj}
\end{equation}
represents the sub-activation-vector of the given $\mathbf{x}_{k}^{\scalebox{0.66}{$(\mathbf{S}_{i})$}}$ and its value is related to the relationship between subset of all labels $\mathbf{S}_{i}$ and $\mathbf{T}_{j}$. Note that the superscription, $\mathbf{S}_{i}$, doesn't imply an additional index of $\mathbf{z}_{k}^{\scalebox{0.66}{$(\mathbf{S}_{i},\mathbf{T}_{j})$}}$. Instead, it represents how does the $\mathbf{z}_{k}^{\scalebox{0.66}{$(\mathbf{S}_{i},\mathbf{T}_{j})$}}$ being treated based on the relationship between $\mathbf{S}_{i}$ and $\mathbf{T}_{j}$.

\paragraph{Decoder:} The decoder performs a partially connected feed-forward neural network \cite{Kang:05}, in which, the neurons are connected differently based on their input types, \textit{e.g.},
\begin{align}
&\mathbf{y}_{k}^{\scalebox{0.66}{$(\mathbf{S}_{i})$}}=a_{\text{decoder}}\left(\sum_{j=1}^{Q^{\scalebox{0.66}{$(\mathbf{S}_{i})$}}}\mathbf{W}^{\scalebox{0.66}{$(\mathbf{T}_{j})$}}\mathbf{z}_{k}^{\scalebox{0.66}{$(\mathbf{S}_{i},\mathbf{T}_{j})$}}+\mathbf{b}_{\text{decoder}}\right),\text{ where}\nonumber\\
&\lbrace\mathbf{T}_{j}|\mathbf{S}_{i}\cap\mathbf{T}_{j}\not\equiv\emptyset,\forall j\rbrace_{Q^{\scalebox{0.66}{$(\mathbf{S}_{i})$}}}.
\end{align}
$a_{\text{decoder}(\cdot)}$ is the activation for the decoder;
$\mathbf{W}^{\scalebox{0.66}{$(\mathbf{T}_{j})$}}$ is a subset of all features defined in Eq.~(\ref{eq:W}); 
$\mathbf{b}_{\text{decoder}}\in\mathbb{R}^{M}$ is the bias term for the decoder;
$\mathbf{y}_{k}^{\scalebox{0.66}{$(\mathbf{S}_{i})$}}$ is the output of the decoder, which is reconstruction of the $\mathbf{x}_{k}^{\scalebox{0.66}{$(\mathbf{S}_{i})$}}$ of the encoder defined in Eq.~(\ref{eq:zkSi}); 
$Q^{\scalebox{0.66}{$(\mathbf{S}_{i})$}}=|\lbrace\mathbf{T}_{j}|\mathbf{S}_{i}\cap\mathbf{T}_{j}\not\equiv\emptyset,\forall j\rbrace|$ is the number of all $\mathbf{T}_{j}$ satisfying this condition with the given $\mathbf{S}_{i}$. In other words, $\mathbf{y}_{k}^{\scalebox{0.66}{$(\mathbf{S}_{i})$}}$ is computed using a subset of all possible $\mathbf{T}_{j}$, each of which is a \textit{non-disjoint set} of $\mathbf{S}_{i}$.

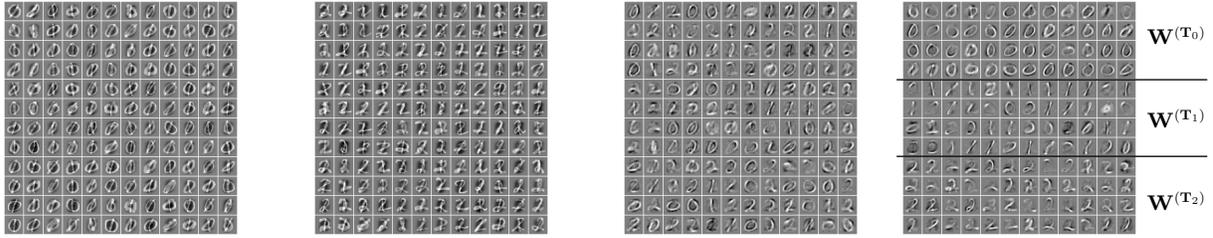
\begin{figure*}%
	\begin{subfigure}[t]{0.25\textwidth}%
		\centering%
		\includegraphics[width=0.75\textwidth]{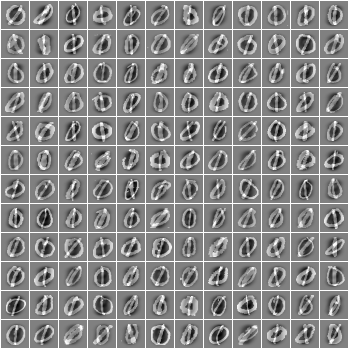}%
		\caption{Mixed digits of `0' and `1'.}%
		\label{fig:mnist_batch_01_response}%
	\end{subfigure}%
	\begin{subfigure}[t]{0.25\textwidth}%
		\centering%
		\includegraphics[width=0.75\textwidth]{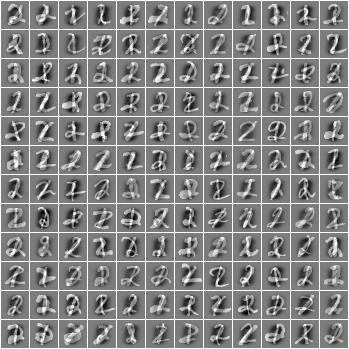}%
		\caption{Mixed digits of `1' and `2'.}%
		\label{fig:mnist_batch_12_response}%
	\end{subfigure}%
	\begin{subfigure}[t]{0.25\textwidth}%
		\centering%
		\includegraphics[width=0.75\textwidth]{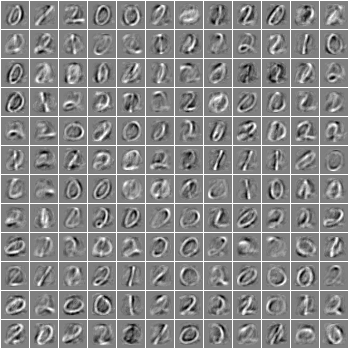}%
		\caption{Conventional AE results.}%
		\label{fig:mnist_ae_weights}%
	\end{subfigure}%
	\begin{subfigure}[t]{0.25\textwidth}
		\begin{tikzpicture}
		\node[inner sep=0pt] (russell) at (0,0)
		{\includegraphics[width=0.75\textwidth]{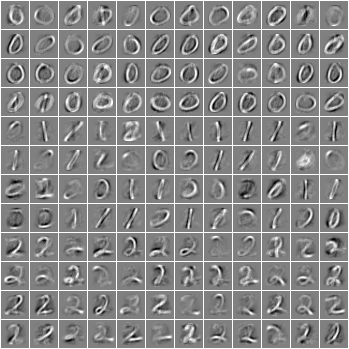}};
		\draw[line width=0.5pt] (-1.63,0.51) -- (2.5,0.51);
		\draw[line width=0.5pt] (-1.63,-0.51) -- (2.5,-0.51);
		\node[text width=3cm] at (3.2,1.1) {\footnotesize $\mathbf{W}^{\scalebox{0.66}{$(\mathbf{T}_{0})$}}$};
		\node[text width=3cm] at (3.2,0) {\footnotesize $\mathbf{W}^{\scalebox{0.66}{$(\mathbf{T}_{1})$}}$};
		\node[text width=3cm] at (3.2,-1.1) {\footnotesize $\mathbf{W}^{\scalebox{0.66}{$(\mathbf{T}_{2})$}}$};
		\end{tikzpicture}
		\caption{XAE features.~~~~~~~~~~~~~~~}
		\label{fig:mnist012_weights}
	\end{subfigure}
	\caption{A toy example using MNIST.}
	\label{fig:mnist}
\end{figure*}%

The cost function of an XAE is defined as
\begin{align}
&\mathcal{J}_{\text{XAE}}=\nonumber\\
&\frac{1}{2|\mathbf{X}^{\scalebox{0.66}{$(\mathbf{S}_{i})$}}|}\sum_{i,k}||\mathbf{x}_{k}^{\scalebox{0.66}{$(\mathbf{S}_{i})$}}-\mathbf{y}_{k}^{\scalebox{0.66}{$(\mathbf{S}_{i})$}}||_{2}^{2}+\frac{\lambda}{2}||\mathbf{W}||_{F}^{2}\label{eqn:cost_part1}\\
&+\alpha v_{\scalebox{0.66}{$\mathbf{S}_{i},\mathbf{T}_{j}$}}\sum_{i,j}\left(\frac{1}{n_{j}}\sum_{m=1}^{n_{j}}\text{KL}\left(\rho||\rho_{m}^{\scalebox{0.66}{$(\mathbf{S}_{i},\mathbf{T}_{j})$}}\right)\right)\label{eqn:cost_part2}\\
&+\frac{\beta\bar{v}_{\scalebox{0.66}{$\mathbf{S}_{i},\mathbf{T}_{j}$}}}{2}\sum_{i,j}\left(\frac{1}{n_{j}|\mathbf{X}^{\scalebox{0.66}{$(\mathbf{S}_{i})$}}|}\sum_{k=1}^{|\scalebox{0.66}{$\mathbf{X}^{\scalebox{0.66}{$(\mathbf{S}_{i})$}}$}|}\sum_{m=1}^{n_{j}}H_{a}\left(\tau_{0,\sigma}||z_{k,m}^{\scalebox{0.66}{$(\mathbf{S}_{i},\mathbf{T}_{j})$}}\right)\right)^{2}\label{eqn:cost_part3}\\
&+\frac{\gamma\bar{v}_{\scalebox{0.66}{$\mathbf{S}_{i},\mathbf{T}_{j}$}}}{2}\sum_{i,j}\left(\frac{1}{|\mathbf{X}^{\scalebox{0.66}{$(\mathbf{S}_{i})$}}|}\sum_{k=1}^{|\scalebox{0.66}{$\mathbf{X}^{\scalebox{0.66}{$(\mathbf{S}_{i})$}}$}|}\left(\mathbf{z}_{k}^{\scalebox{0.66}{$(\mathbf{S}_{i},\mathbf{T}_{j})$}}\right)\left(\mathbf{z}_{k}^{\scalebox{0.66}{$(\mathbf{S}_{i},\mathbf{T}_{j})$}}\right)^{T}-b\mathbf{I}\right)^{2},\label{eqn:cost_part4}
\end{align}
where $\alpha$, $\beta$ and $\gamma$ are parameters controlling the weights of sparseness and exclusiveness; 
\begin{equation}
\rho_{m}^{\scalebox{0.66}{$(\mathbf{S}_{i},\mathbf{T}_{j})$}}=\frac{1}{|\mathbf{X}^{\scalebox{0.66}{$(\mathbf{S}_{i})$}}|}\sum_{k=1}^{|\scalebox{0.66}{$\mathbf{X}^{\scalebox{0.66}{$(\mathbf{S}_{i})$}}$}|}
z_{k,m}^{\scalebox{0.66}{$(\mathbf{S}_{i},\mathbf{T}_{j})$}}
\end{equation}
%
%
is the sparsity measurement of the $m^{th}$ element of the sub-activation-vector $\mathbf{z}_{k}^{\scalebox{0.66}{$(\mathbf{S}_{i},\mathbf{T}_{j})$}}$ in Eq.~(\ref{eq:zkSiTj}); $\text{KL}(\rho||\cdot)$ is the measurement of KL divergence, which requires a parameter $\rho$, indicating the level of sparsity \cite{Ng}; and 
\begin{equation}
H_{a}\left(\tau_{0,\sigma}||z_{k,m}^{\scalebox{0.66}{$(\mathbf{S}_{i},\mathbf{T}_{j})$}}\right)=G_{a}\left(z_{k,m}^{\scalebox{0.66}{$(\mathbf{S}_{i},\mathbf{T}_{j})$}}\right)-G_{a}(\tau_{0,\sigma})
\end{equation}
measures the Gaussianity of  the given $z_{k,m}^{\scalebox{0.66}{$(\mathbf{S}_{i},\mathbf{T}_{j})$}}$ so that the response of an excluded feature approximates to a Gaussian random variable defined by $\tau_{0,\sigma}$. $G(\cdot)$ is a non-quadratic function \cite{Hyvarinen:00:411}, \textit{e.g.},
\begin{equation}
G_{a}(u)=\left\lbrace%
\begin{array}{l}%
\frac{1}{a}\log\cosh(au)\\
\frac{-1}{a}exp\left(\frac{-1}{2}au^2\right)
\end{array}
\right.
\end{equation}
where $a$ is a scale parameter which is correlated to the parameter $b$ in Eq.~(\ref{eqn:cost_part4}). $\tau_{0,\sigma}$ is a zero-mean Gaussian random variable with a small standard deviation of $\sigma$. Finally, $v_{\scalebox{0.66}{$\mathbf{S}_{i},\mathbf{T}_{j}$}}$ represents a semaphore determined by the relationship between $\mathbf{S}_{i}$ and $\mathbf{T}_{j}$:
\begin{equation}
v_{\scalebox{0.66}{$\mathbf{S}_{i},\mathbf{T}_{j}$}}=\left\lbrace
\arraycolsep=1.4pt
\begin{array}{ll}
1,&\text{if }\mathbf{S}_{i}\cap\mathbf{T}_{j}\not\equiv\emptyset\\
0,&\text{otherwise}
\end{array}
\right.,\bar{v}_{\scalebox{0.66}{$\mathbf{S}_{i},\mathbf{T}_{j}$}}=1-v_{\scalebox{0.66}{$\mathbf{S}_{i},\mathbf{T}_{j}$}}.
\end{equation}

In a nutshell, the cost function can be interpreted as the following: Eq.~(\ref{eqn:cost_part1}) performs the same function as a conventional AE. If $\mathbf{S}_{i}\cap\mathbf{T}_{j}\not\equiv\emptyset$, Eq.~(\ref{eqn:cost_part2}) expects the sparsification of $\mathbf{z}_{k}^{\scalebox{0.66}{$(\mathbf{S}_{i},\mathbf{T}_{j})$}}$; on the other hand, Eq.~(\ref{eqn:cost_part3}) and (\ref{eqn:cost_part4}) act like the counter part of Eq.~(\ref{eqn:cost_part2}). That is, if $\mathbf{S}_{i}\cap\mathbf{T}_{j}\equiv\emptyset$, Eq.~(\ref{eqn:cost_part3}) and (\ref{eqn:cost_part4}), result Gaussanization, where Eq.~(\ref{eqn:cost_part3}) Gaussianize each $z_{k,m}^{\scalebox{0.66}{$(\mathbf{S}_{i},\mathbf{T}_{j})$}}$ and Eq.~(\ref{eqn:cost_part4}) minimize the correlation between each $z_{k,m}^{\scalebox{0.66}{$(\mathbf{S}_{i},\mathbf{T}_{j})$}}$ and others. Hence, if we wish that a $\mathbf{W}^{\scalebox{0.66}{$(\mathbf{T}_{j})$}}$ learns the features from a $\mathbf{X}^{\scalebox{0.66}{$(\mathbf{S}_{i})$}}$, sparsify $z_{k,m}^{\scalebox{0.66}{$(\mathbf{S}_{i},\mathbf{T}_{j})$}}$, otherwise, Gaussianize $z_{k,m}^{\scalebox{0.66}{$(\mathbf{S}_{i},\mathbf{T}_{j})$}}$.

\subsection{AE v.s. XAE}

One may be confused by the idea that the XAE is equivalent to a parallel combination of two or more conventional AEs, which is not true for the following reasons: 1) a conventional AE DOES NOT exclude the features co-existing in other classes of sub-dataset; 2) the mutual features identified by XAE  exist in all corresponding classes, while for an conventional AE, that is not necessary true; and 3) the only way making XAEs emulating the behavior of conventional AEs is setting proper $v_{\scalebox{0.66}{$\mathbf{S}_{i},\mathbf{T}_{j}$}}$ and disabling Eqs.~(\ref{eqn:cost_part3}) and (\ref{eqn:cost_part4}), due to the fact that both the mutual and exclusive features may or may not co-exist in some or more classes of the dataset.

\subsection{Toy Example using MNIST}
\label{sec:toyexample}

To understand the further details of the proposed XAE, we started from a simple example of MNIST handwritten digit mixtures \cite{MNISTHandwrittenDigitDatabase} to validate the capability of mutual/exclusive feature extraction using XAE. In order to simplify the situation, we used only handwritten digits of `0', `1' and `2'. First,we  allocated two sub-datasets, say $\mathbf{X}^{\scalebox{0.66}{$(\mathbf{S}_{0})$}}$, contained `0' and `1' while $\mathbf{X}^{\scalebox{0.66}{$(\mathbf{S}_{1})$}}$ possessed `1' and `2'. Note that there was no data instance duplicated in both the sub-datasets. Next, we used the digits as the labels defining the label set
\begin{equation}
\mathbf{L}=\left\lbrace\text{`0'},\text{`1'},\text{`2'}\right\rbrace.
\end{equation}
Subsequently, defined
\begin{equation}
\mathbf{S}_{0}=\left\lbrace\text{`0'},\text{`1'}\right\rbrace\text{ and }\mathbf{S}_{1}=\left\lbrace\text{`1'},\text{`2'}\right\rbrace
\end{equation}
for $\mathbf{X}^{\scalebox{0.66}{$(\mathbf{S}_{0})$}}$ and $\mathbf{X}^{\scalebox{0.66}{$(\mathbf{S}_{1})$}}$, respectively. Here $\mathbf{S}_{i}\forall i$ represented the \textit{priori knowledge} of $\mathbf{X}^{\scalebox{0.66}{$(\mathbf{S}_{i})$}}\forall i$. The examples are shown in Fig.~\ref{fig:mnist_batch_01_response} and Fig.~\ref{fig:mnist_batch_12_response}, respectively.

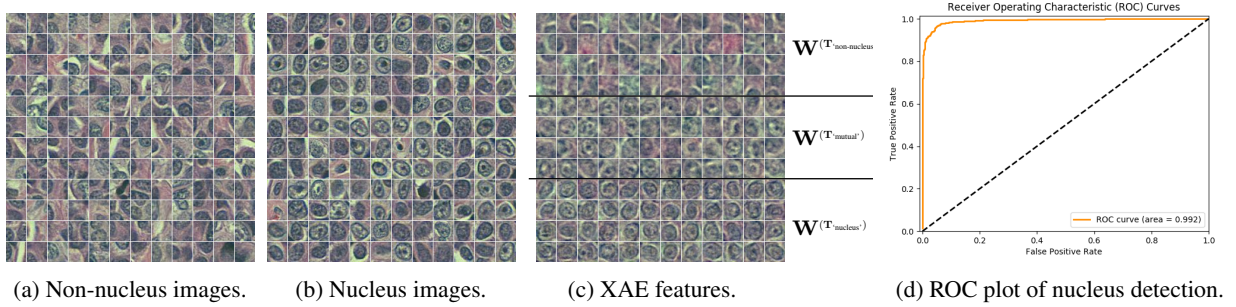
\begin{figure*}
	\begin{center}
		\begin{subfigure}[t]{0.2\textwidth}
			\includegraphics[width=\textwidth]{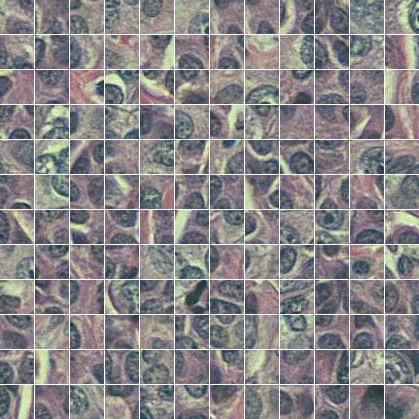}
			\caption{Non-nucleus images.}
			\label{fig:nucleus_images}
		\end{subfigure}%
		~
		\begin{subfigure}[t]{0.2\textwidth}
			\includegraphics[width=\textwidth]{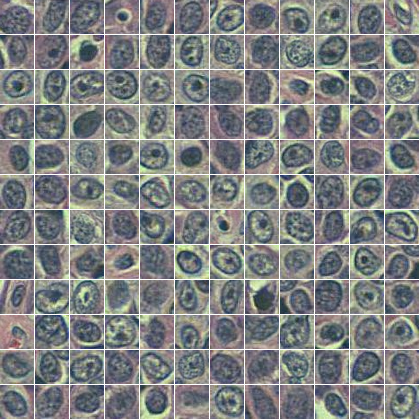}
			\caption{Nucleus images.}
			\label{fig:non_nucleus_images}
		\end{subfigure}%
		~
		\begin{subfigure}[t]{0.2\textwidth}
			\begin{tikzpicture}
			\node[inner sep=0pt] (russell) at (0,0)
			{\includegraphics[width=\textwidth]{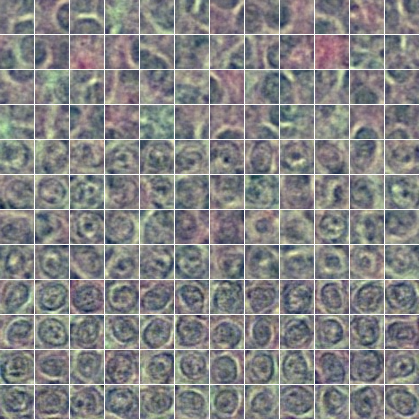}};
			\draw[line width=0.5pt] (-1.75,0.55) -- (3,0.55);
			\draw[line width=0.5pt] (-1.75,-0.55) -- (3,-0.55);
			\node[text width=3cm] at (3.25,1.2) {\footnotesize $\mathbf{W}^{\scalebox{0.5}{$(\mathbf{T}_{\text{`non-nucleus'}})$}}$};
			\node[text width=3cm] at (3.25,0) {\footnotesize $\mathbf{W}^{\scalebox{0.5}{$(\mathbf{T}_{\text{`mutual'}})$}}$};
			\node[text width=3cm] at (3.25,-1.2) {\footnotesize $\mathbf{W}^{\scalebox{0.5}{$(\mathbf{T}_{\text{`nucleus'}})$}}$};
			\end{tikzpicture}
			\caption{XAE features.}
			\label{fig:nucleus_weights}
		\end{subfigure}
		~~~~~~~~~~~~~
		\begin{subfigure}[t]{0.3\textwidth}
			\includegraphics[width=\linewidth]{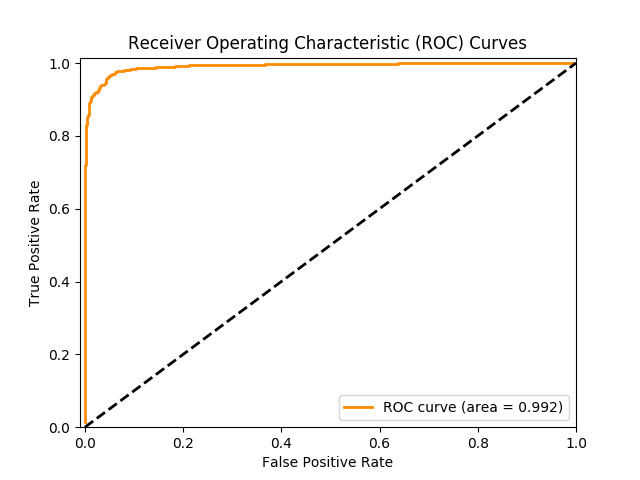}
			\caption{ROC plot of nucleus detection.}
			\label{fig:detection_rocplot}
		\end{subfigure}\\
		
	\end{center}
	\caption{The corresponding images of nucleus detection and the ROC plot.}
	\label{fig:nucleus_detection}
\end{figure*}%

\begin{table*}
	
	\begin{center}
		\scalebox{0.8}{\begin{tabular}{|c|c|c|c|c|}
				\hline
				Model                    & F-score (\%) & Precision (\%) & Recall (\%) & Average Precision (\%) \\ \hline
				LoG of Nuclear Channel + CNN \cite{Khoshdeli:17:105} &         72.22         &        74.33        &        69.78         &           N/A            \\ \hline
				Nuclear Channel + CNN \cite{Khoshdeli:17:105}     &         66.99         &        71.51        &        63.01         &           N/A            \\ \hline
				RGB + CNN \cite{Khoshdeli:17:105}           &         53.61         &        88.94        &        38.36         &           N/A            \\ \hline\hline
				Expectation-Maximum                  &         70.84         &        66.38        &        78.73         &          54.78           \\ \hline
				Blue Ratio Thresholding                &         71.04         &        84.40        &        61.87         &          49.25           \\ \hline
				Color Deconvolution                  &         75.61         &        77.75        &        75.80         &          59.48           \\ \hline
				Softmax Classifier \cite{Xu:16:119}          &         79.56         &        78.05        &        78.39         &          68.29           \\ \hline
				1-Layer AE + Softmax \cite{Xu:16:119}         &         79.85         &        83.51        &        77.40         &          68.09           \\ \hline
				1-Layer Sparse AE + Softmax \cite{Xu:16:119}     &         80.68         &        84.52        &        77.93         &          70.51           \\ \hline
				Stacked AE + Softmax \cite{Xu:16:119}         &         83.12         &        83.71        &        82.98         &          78.29           \\ \hline
				3-Layer Sparse AE + Softmax \cite{Xu:16:119}     &         82.26         &        88.35        &        77.15         &          73.48           \\ \hline
				CNN + Softmax \cite{Xu:16:119}            &         82.01         &        88.28        &        77.60         &          73.65           \\ \hline
				Stacked Sparse AE + Softmax \cite{Xu:16:119}     &         84.49         &        88.84        &        82.85         &          78.83           \\ \hline
				\textbf{1-Layer XAE + Softmax}                 &    \textbf{94.82}     &   \textbf{95.18}    &    \textbf{94.48}    &      \textbf{87.15}      \\ \hline
				\textbf{1-Layer XAE + FCN}                   &    \textbf{96.64}     &   \textbf{96.15}    &    \textbf{97.15}    &      \textbf{90.83}      \\ \hline
		\end{tabular}}
	\end{center}
	\caption{Comparisons with conventional computer vision approaches, the methods reported in \cite{Xu:16:119, Khoshdeli:17:105}, the proposed XAE+FCN and XAE+Softmax based on 10-cross-validation. Note that in the table, most of the results were based on the same dataset, except the methods reported in \cite{Khoshdeli:17:105}.}
	\label{tbl:nuclei_detection}
\end{table*}

We were interested on the features of each handwritten digits of `0', `1' and `2', Thus, we defined
\begin{equation}
\mathbf{T}_{0}=\left\lbrace\text{`0'}\right\rbrace,
\mathbf{T}_{1}=\left\lbrace\text{`1'}\right\rbrace\text{ and }
\mathbf{T}_{2}=\left\lbrace\text{`2'}\right\rbrace
\end{equation}
for
$\mathbf{W}^{\scalebox{0.66}{$(\mathbf{T}_{0})$}}$
$\mathbf{W}^{\scalebox{0.66}{$(\mathbf{T}_{1})$}}$, and
$\mathbf{W}^{\scalebox{0.66}{$(\mathbf{T}_{2})$}}$,
respectively. $T_{i}\forall i$ defined the \textit{desired arrangement} for each $\mathbf{W}^{\scalebox{0.66}{$(\mathbf{T}_{i})$}}\forall i$. This configuration is much similar to the plot showing in Fig.~\ref{fig:xae}. 

For the comparison in between AE and XAE, we performed conventional AE on the union of the two sub-datasets, $\mathbf{X}^{\scalebox{0.66}{$(\mathbf{S}_{0})$}}\cup\mathbf{X}^{\scalebox{0.66}{$(\mathbf{S}_{1})$}}$, in order to understand how the conventional AE performs feature selection in this dataset. an obtained result is showing in Fig.~\ref{fig:mnist_ae_weights}. One can find that the conventional AE performed as what we expected, extracted the key features from the given datasets. Each of these features is orthogonal to another and the all features span across the whole data space. 
\newpage
Finally, we performed XAE with the given sub-datasets, $\mathbf{X}^{\scalebox{0.66}{$(\mathbf{S}_{0})$}}$ and $\mathbf{X}^{\scalebox{0.66}{$(\mathbf{S}_{1})$}}$, using the parameters: $\lambda=1$, $\beta=1$, $\gamma=1$, $\text{learning rate }\eta=0.01$, $a_\text{encoder}=\text{sigmoid}$ and $a_\text{decoder}=\text{linear}$, we obtained the results showing in Fig.~\ref{fig:mnist012_weights}. This result suggested that even there was no explicit label for each instance of handwritten digits `0' and `1' in $\mathbf{X}^{(\mathbf{S}_{0})}$ (as well as `1' and `2' in $\mathbf{X}^{(\mathbf{S}_{1})}$), XAE was managed to identify their exclusive features of `0' and `2', also their mutual features of `1' with the desired feature space partitioning. Note that since the handwritten digits `0' and `2' shared some mutual features, which are also presented in Fig.~\ref{fig:mnist012_weights}. In this case, XAE is able to discover the fact that: 1) feature set $\mathbf{W}^{\scalebox{0.66}{$(\mathbf{T}_{0})$}}$ possesses the key features of hand written digits of `0', $\mathbf{W}^{\scalebox{0.66}{$(\mathbf{T}_{0})$}}$ is also the features set which is not correlated to hand written digits of `2' and there is the similar situation in between $\mathbf{W}^{\scalebox{0.66}{$(\mathbf{T}_{2})$}}$ and hand written digits of '0'; and 2) $\mathbf{W}^{\scalebox{0.66}{$(\mathbf{T}_{1})$}}$ is a special feature set as it exists in both of sub-dataset. 

In order to validate the improvement based on XAE, we performed classification on hand written digits of `1', `2' and `3' based on the training dataset described above. In the result, the mean F-score (10-fold cross-validation) of XAE+NN was $98.2\%$ while AE+NN was $97.7\%$. Note that in this experiment, the architectures of AE+NN and XAE+NN were almost identical, except XAE took the data labels into account. The result suggested great potential of XAE for improving the classification performance.





\begin{figure*}%
	
	\begin{center}
	\begin{subfigure}[t]{0.2\textwidth}%
		\centering%
		\includegraphics[width=0.9\textwidth]{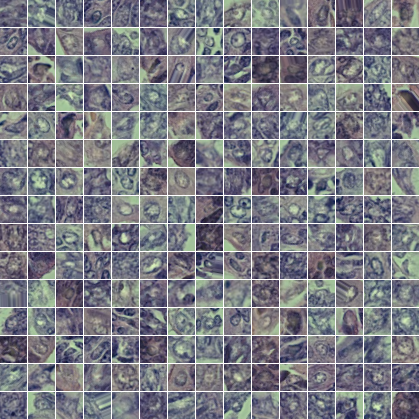}%
		\caption{Epithelial.}%
		\label{fig:classification_epthi}%
	\end{subfigure}%
	\begin{subfigure}[t]{0.2\textwidth}%
		\centering%
		\includegraphics[width=0.9\textwidth]{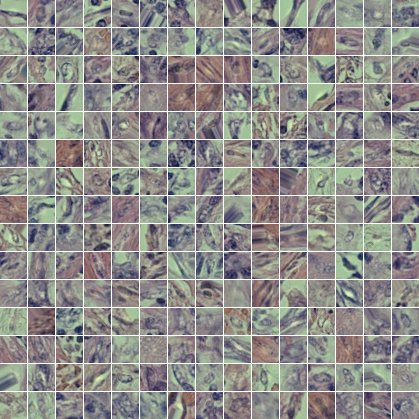}%
		\caption{Inflammatory.}%
		\label{fig:classification_inflam}%
	\end{subfigure}%
	\begin{subfigure}[t]{0.2\textwidth}%
		\centering%
		\includegraphics[width=0.9\textwidth]{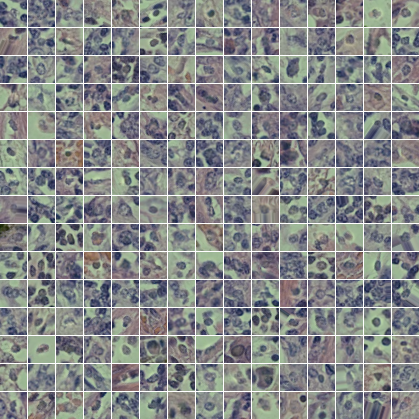}%
		\caption{Fibroblast.}%
		\label{fig:classification_fibro}%
	\end{subfigure}%
	\begin{subfigure}[t]{0.2\textwidth}%
		\centering%
		\includegraphics[width=0.9\textwidth]{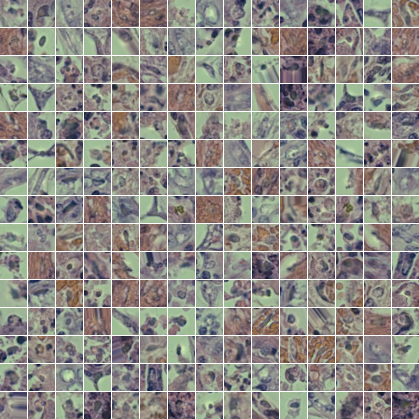}%
		\caption{Miscellaneous.}%
		\label{fig:classification_misc}%
	\end{subfigure}%
	\begin{subfigure}[t]{0.2\textwidth}
		\begin{tikzpicture}
		\node[inner sep=0pt] (russell) at (0,0)
		{\includegraphics[width=0.9\textwidth]{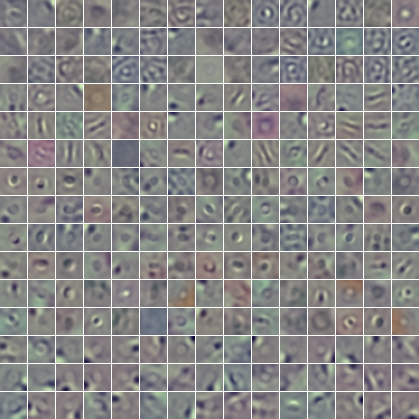}};
		\draw[line width=0.5pt] (-1.57,0.89) -- (2.75,0.89);
		\draw[line width=0.5pt] (-1.57,0.3) -- (2.75,0.3);
		\draw[line width=0.5pt] (-1.57,-0.3) -- (2.75,-0.3);
		\draw[line width=0.5pt] (-1.57,-0.89) -- (2.75,-0.89);
		\node[text width=3cm] at (3.1,1.25) {\footnotesize $\mathbf{W}^{\scalebox{0.66}{$(\mathbf{T}_{\text{`epith.'}})$}}$};
		\node[text width=3cm] at (3.1,0.625) {\footnotesize $\mathbf{W}^{\scalebox{0.66}{$(\mathbf{T}_{\text{`inflam.'}})$}}$};
		\node[text width=3cm] at (3.1,0) {\footnotesize $\mathbf{W}^{\scalebox{0.66}{$(\mathbf{T}_{\text{`fibro.'}})$}}$};
		\node[text width=3cm] at (3.1,-0.625) {\footnotesize $\mathbf{W}^{\scalebox{0.66}{$(\mathbf{T}_{\text{`misc.'}})$}}$};
		\node[text width=3cm] at (3.1,-1.25) {\footnotesize $\mathbf{W}^{\scalebox{0.66}{$(\mathbf{T}_{\text{`mutual'}})$}}$};
		\end{tikzpicture}
		\caption{XAE features.~~~~~}
		\label{fig:classification_weights}
	\end{subfigure}
\end{center}
\caption{Some examples of training patterns and XAE features obtained in the experiments. Note that for the sake of better displaying the weights here, only 225 are showed. In the experiment, there were 1280 weights used in total.}
\label{fig:classification}
\end{figure*}
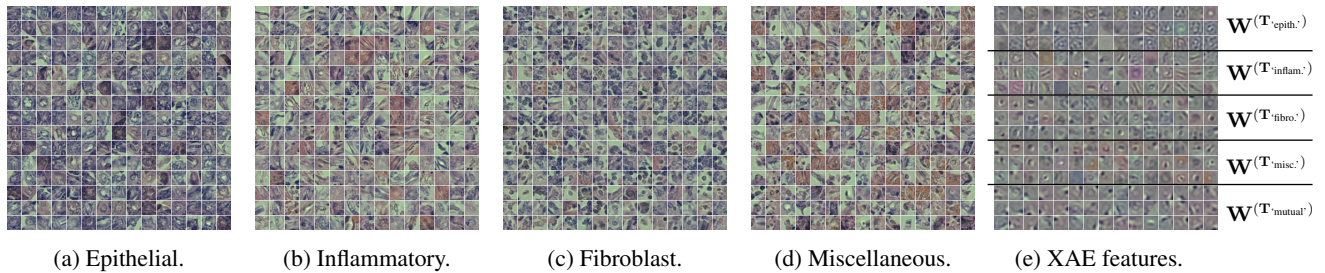%

\begin{figure}[t]
	\begin{center}
		\includegraphics[width=1.0\linewidth]{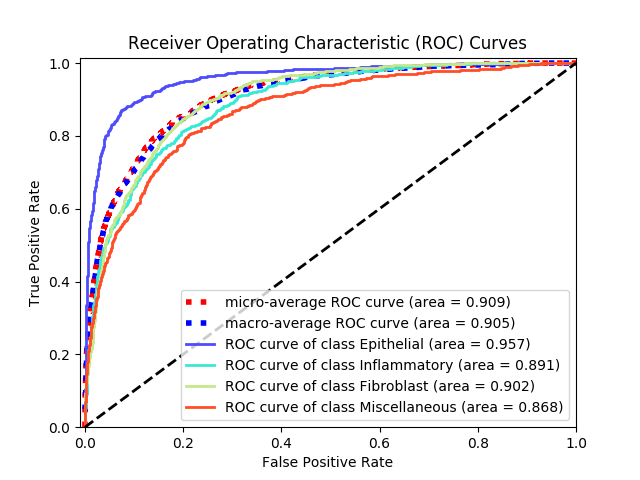}
	\end{center}
	\caption{The ROC of the XAE+FCN for nucleus classification.}
	\label{fig:classification_roc}
\end{figure}



\begin{table}
	\begin{center}
		\scalebox{0.8}{\begin{tabular}{|c||c|c|c|c|c|}
				\hline
				~    & Nuclei & Non-nuclei &  Accuracy (\%)  \\ \hline\hline
				Nuclei &   2386   &   38   & 98.43 \\ \hline
				Non-nuclei &   79    &   697  & 89.82 \\ \hline
		\end{tabular}}
	\end{center}
	\caption{Confusion matrix of nucleus detection (total: 3200).}
	\label{tbl:detection_conf_matrix}
\end{table}

\begin{table}
	\begin{center}
	\scalebox{0.8}{\begin{tabular}{|c||c|c|c|c|c|}
	\hline
	~    & Epith. & Inflam. & Fibro. & Misc. &  Accuracy (\%)  \\ \hline\hline
	Epith. &   900   &   76    &   52   &  32   & 84.91 \\ \hline
	Inflam. &   108    &   629   &   118   &  93   & 66.35 \\ \hline
	Fibro.  &   47    &   88    &  737   &  185   & 69.73 \\ \hline
	Misc.  &    46    &   173    &   182   &  630  & 61.11 \\ \hline
\end{tabular}}
	\end{center}
	\caption{Confusion matrix of nucleus classification (total: 4096)}
	\label{tbl:classification_conf}
\end{table}

%
%

\begin{table}
	\begin{center}
		\scalebox{0.8}{\begin{tabular}{|c||c|c|}
			\hline
			\multirow{ 2}{*}{Method} & Weighted & Multi-class \\ & Avg. F-score (\%) & AUC (\%) \\  \hline\hline
			CRImage \cite{Yuan:12:157ra143} & 48.8 & 68.4 \\ \hline 		
			Softmax CNN + NEP \cite{Sirinukunwattana:16:1196} & 78.4 & 91.7 \\ \hline 
			Superpixel Descriptor \cite{Sirinukunwattana:15:94200S} & 68.7 & 85.3 \\ \hline 		
			Softmax CNN + SSPP \cite{Sirinukunwattana:16:1196} & 74.8 & 89.3 \\ \hline 
			\textbf{1-Layer XAE + FCN} & \textbf{70.4} & \textbf{90.7} \\ \hline
		\end{tabular}}
	\end{center}
	\caption{Comparative results of nucleus classification.}
	\label{tbl:classification_comparison}
\end{table}

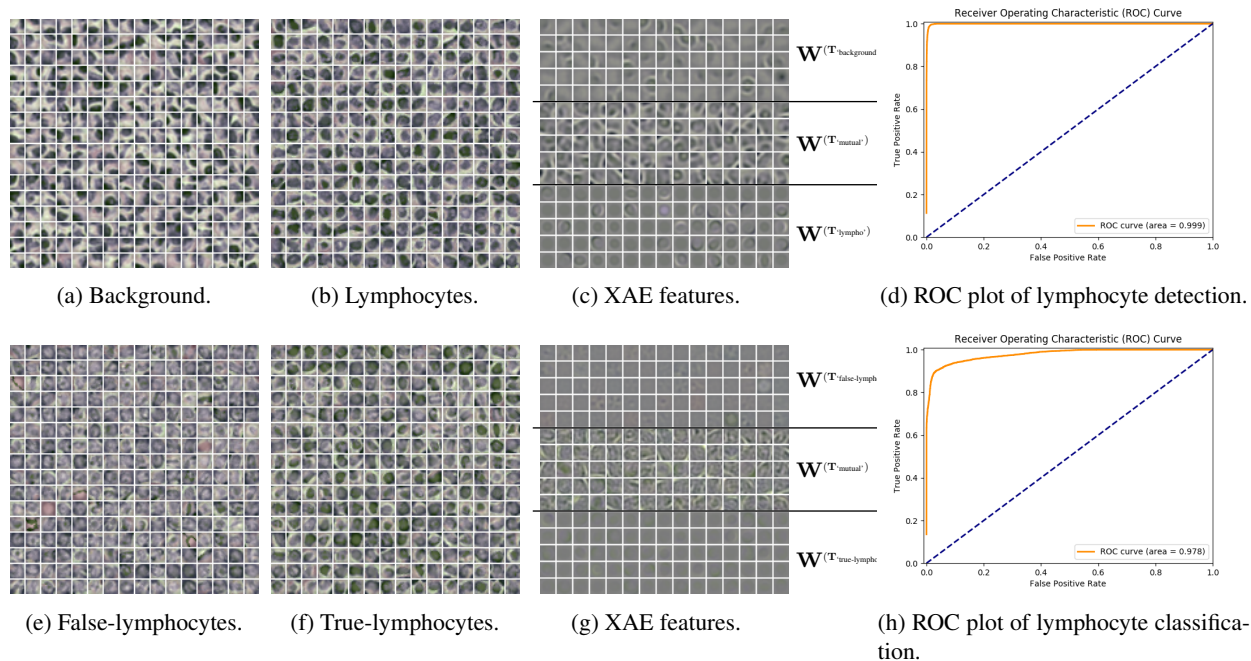
\begin{figure*}
	\begin{center}
		\begin{subfigure}[t]{0.2\textwidth}
			\includegraphics[width=\textwidth]{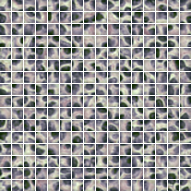}
			\caption{Background.}
			\label{fig:lymphocyte_detection_backgrd}
		\end{subfigure}%
		~
		\begin{subfigure}[t]{0.2\textwidth}
			\includegraphics[width=\textwidth]{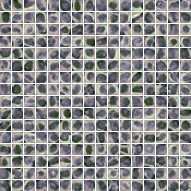}
			\caption{Lymphocytes.}
			\label{fig:lymphocyte_detection_lympho}
		\end{subfigure}%
		~
		\begin{subfigure}[t]{0.2\textwidth}
			\begin{tikzpicture}
			\node[inner sep=0pt] (russell) at (0,0)
			{\includegraphics[width=\textwidth]{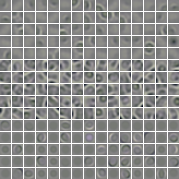}};
			\draw[line width=0.5pt] (-1.75,0.553) -- (3,0.553);
			\draw[line width=0.5pt] (-1.75,-0.553) -- (3,-0.553);
			\node[text width=3cm] at (3.25,1.2) {\footnotesize $\mathbf{W}^{\scalebox{0.5}{$(\mathbf{T}_{\text{`background'}})$}}$};
			\node[text width=3cm] at (3.25,0) {\footnotesize $\mathbf{W}^{\scalebox{0.5}{$(\mathbf{T}_{\text{`mutual'}})$}}$};
			\node[text width=3cm] at (3.25,-1.2) {\footnotesize $\mathbf{W}^{\scalebox{0.5}{$(\mathbf{T}_{\text{`lympho'}})$}}$};
			\end{tikzpicture}
			\caption{XAE features.}
			\label{fig:lymphocyte_detection_weights}
		\end{subfigure}
		~~~~~~~~~~~~~
		\begin{subfigure}[t]{0.3\textwidth}
			\includegraphics[width=\linewidth]{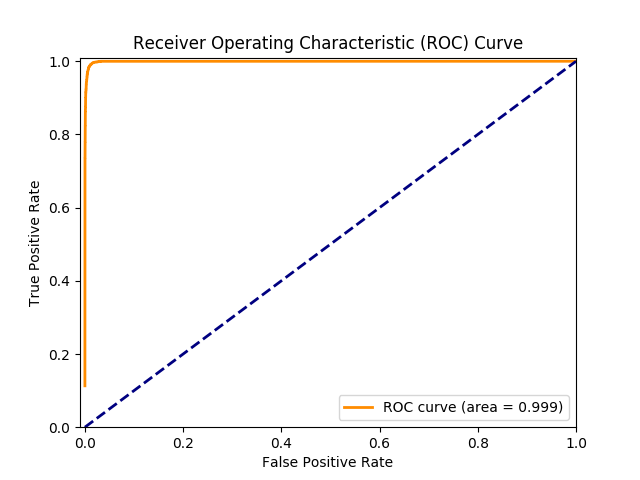}
			\caption{ROC plot of lymphocyte detection.}
			\label{fig:lymphocyte_detection_roc}
		\end{subfigure}\\
		\begin{subfigure}[t]{0.2\textwidth}
			\includegraphics[width=\textwidth]{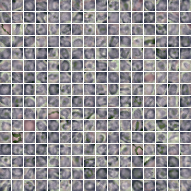}
			\caption{False-lymphocytes.}
			\label{fig:lymphocyte_classification_false}
		\end{subfigure}%
		~
		\begin{subfigure}[t]{0.2\textwidth}
			\includegraphics[width=\textwidth]{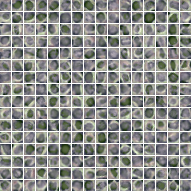}
			\caption{True-lymphocytes.}
			\label{fig:lymphocyte_classification_lympho}
		\end{subfigure}%
		~		
		\begin{subfigure}[t]{0.2\textwidth}
			\begin{tikzpicture}
			\node[inner sep=0pt] (russell) at (0,0)
			{\includegraphics[width=\textwidth]{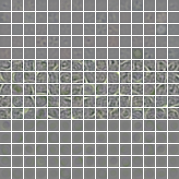}};
			\draw[line width=0.5pt] (-1.75,0.553) -- (3,0.553);
			\draw[line width=0.5pt] (-1.75,-0.553) -- (3,-0.553);
			\node[text width=3cm] at (3.25,1.2) {\footnotesize $\mathbf{W}^{\scalebox{0.5}{$(\mathbf{T}_{\text{`false-lympho'}})$}}$};
			\node[text width=3cm] at (3.25,0) {\footnotesize $\mathbf{W}^{\scalebox{0.5}{$(\mathbf{T}_{\text{`mutual'}})$}}$};
			\node[text width=3cm] at (3.25,-1.2) {\footnotesize $\mathbf{W}^{\scalebox{0.5}{$(\mathbf{T}_{\text{`true-lympho'}})$}}$};
			\end{tikzpicture}
			\caption{XAE features.}
			\label{fig:lymphocyte_classification_weights}
		\end{subfigure}
	~~~~~~~~~~~~~
	\begin{subfigure}[t]{0.3\textwidth}
		\includegraphics[width=\linewidth]{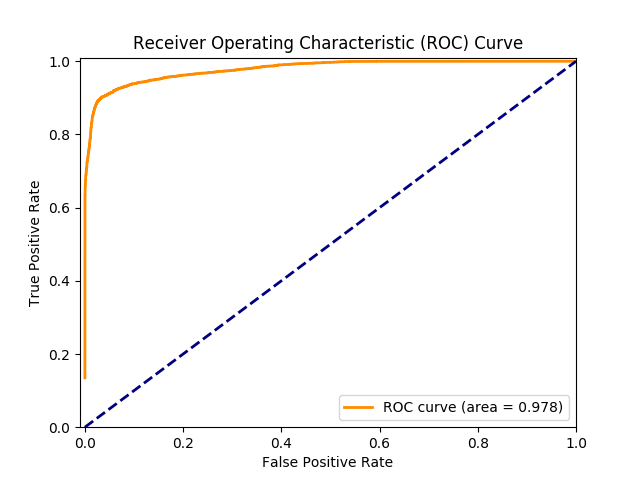}
		\caption{ROC plot of lymphocyte classification.}
		\label{fig:lymphocyte_classification_roc}
	\end{subfigure}
	\end{center}
	\caption{The corresponding images of lymphocyte detection/classification and the corresponding ROC plots: (a)-(d) lymphocyte detection; (e)-(h) lymphocyte classification.}
	\label{fig:lymphocyte}
\end{figure*}%
\begin{figure*}
	\begin{center}
		\begin{subfigure}{0.16\textwidth}
			\includegraphics[width=\textwidth]{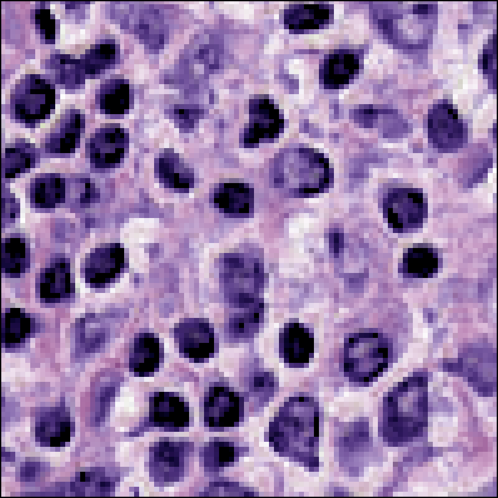}
			\caption{}
		\end{subfigure}~%
		\begin{subfigure}{0.16\textwidth}
			\includegraphics[width=\textwidth]{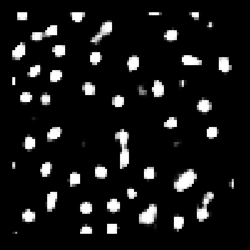}
			\caption{}
		\end{subfigure}~%
		\begin{subfigure}{0.16\textwidth}
			\includegraphics[width=\textwidth]{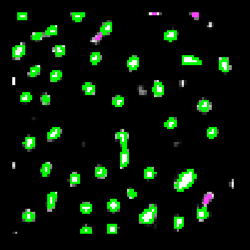}
			\caption{}
		\end{subfigure}~%
		\begin{subfigure}{0.16\textwidth}
			\includegraphics[width=\textwidth]{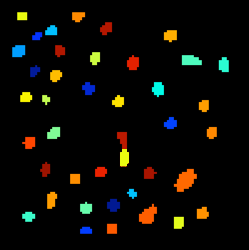}
			\caption{}
		\end{subfigure}~%
		\begin{subfigure}{0.16\textwidth}
			\includegraphics[width=\textwidth]{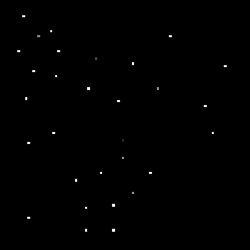}
			\caption{}
		\end{subfigure}~%
		\begin{subfigure}{0.16\textwidth}
			\includegraphics[width=\textwidth]{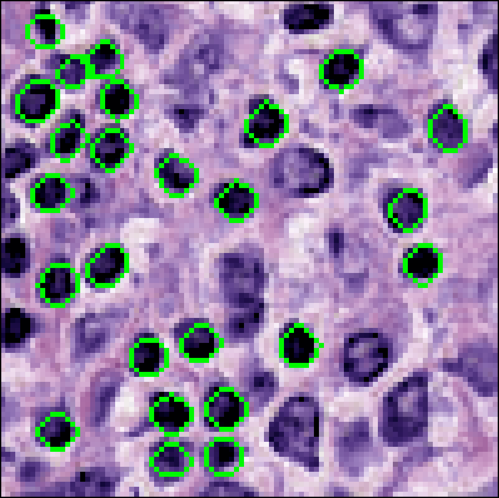}
			\caption{}
		\end{subfigure}%
	\end{center}
	\caption{The procedure of lymphocyte detection/classification: (a) the input; (b) a probability map obtained from XAE-based lymphocyte detection, where the intensity represents the probability that a pixel belongs to a lymphocyte; (c) performing general object detection on the probability map obtained from step (b); (d) performing general object segmentation on the result of object detection, where each object is either a lymphocyte or a false-positive; (e) lymphocyte classification based on XAE, where the intensity represents the probability that a pixel represents the center of a lymphocyte; and (f) the final result of lymphocyte segmentation obtained from the integration of steps (d) and (e).}
	\label{fig:lympho_segmentation}
\end{figure*}

\begin{table}
	\begin{subfigure}[t]{0.45\textwidth}
		\begin{center}
			\scalebox{0.8}{\begin{tabular}{|c||c|c|c|c|c|}
				\hline
				~    & ~Background~ & ~Lymphocyte~ &  ~Accuracy (\%)~  \\ \hline\hline
				~Background~ &   14672   &   205   & 98.62 \\ \hline
				~Lymphocyte~ &   182    &   14118  & 98.73 \\ \hline
			\end{tabular}}
		\end{center}
		\caption{Lymphocyte detection (total: 29177)}
		\label{tbl:lymphocyte_detection_conf}
	\end{subfigure}%
	\\[5pt]
	\begin{subfigure}[t]{0.45\textwidth}
		\begin{center}
			\scalebox{0.8}{\begin{tabular}{|c||c|c|c|c|c|}
				\hline
				~    & False-lympho. & True-lympho. &  Accuracy (\%)  \\ \hline\hline
				False-lympho. &   5079   &   465   & 91.61 \\ \hline
				True-lympho. &  366  &   5016  & 93.20  \\ \hline
			\end{tabular}}
		\end{center}
		\caption{Lymphocyte classification (total: 10926)}
		\label{tbl:lymphocyte_classification_conf}
	\end{subfigure}	
	\caption{Confusion matrix of lymphocyte detection and classification used in the integrated lymphocyte segmentation procedure.}
\end{table}

%
%
%


\section{Results}
\label{sec:results}

\subsection{Nucleus Detection}

In this experiment, we used a dataset\footnote{\url{http://engineering.case.edu/centers/ccipd/data/}} containing 28800 of of $34\times 34$ RGB images, each image was either a nucleus or something else on breast cancer H\&E images. Some examples are shown in Fig.~\ref{fig:nucleus_images} and \ref{fig:non_nucleus_images}.

Nucleus detection, in a way, can be considered as a nucleus/non-nucleus classification problem. First, we performed XAE for nuclei/non-nuclei images and obtained the XAE features (see Fig.~\ref{fig:nucleus_weights}). Then, we used Softmax classifier which is the same as in \cite{Xu:16:119}, and obtained F-score of $94.82\%$. The comparison is showing in Table~\ref{tbl:nuclei_detection}. One can see that the proposed XAE+Softmax largely outperformed accuracy of previously developed methods of AE+Softmax. Finally, we improved the performance by using a FCN instead of the Softmax. The accuracy was further improved. The F-score was $96.62\%$. The corresponding receiver operating characteristic (ROC) plot and the confusion matrix are shown in Fig.~\ref{fig:detection_rocplot} and Table~\ref{tbl:detection_conf_matrix}, respectively.

\subsection{Nucleus Classification}


The dataset\footnote{\url{https://warwick.ac.uk/fac/sci/dcs/research/tia/data/crchistolabelednucleihe/}} used in this experiment was obtained from the report proposed by Sirinukunwattana \textit{et al.} \cite{Sirinukunwattana:16:1196}. In the dataset, four categories of nucleus H\&E images were obtained from colorectal cancer biopsies, including epithelial, inflammatory, fibroblast and miscellaneous nucleus types. Some examples are shown in Fig.~\ref{fig:classification_epthi}, \ref{fig:classification_inflam}, \ref{fig:classification_fibro} and \ref{fig:classification_misc}. The size of each image was $27\times 27$ RGB images, and there were a few thousand of images in each category. 

In order to perform the nucleus classification, first, we obtained the XAE features. some obtained XAE features are showing in Fig.~\ref{fig:classification_weights}. Note that for the sake of better displaying the weights on the paper, only 225 are showed. In the experiment, there were 1280 weights used in total.
\newpage
Then, a stacked FCN was used in order to perform the classification. We used 10-fold cross-validation, and obtained the weighted F-score of $87.5\%$. The corresponding confusion matrix and the ROC can be found in Table~\ref{tbl:classification_conf} and Fig.~\ref{fig:classification_roc}, respectively. The result of a further comparison with existing approaches is showing in Table.~\ref{tbl:classification_comparison}.

\subsection{Lymphocyte Segmentation based on Nucleus Detection/Classification}

An automated lymphocyte segmentation approach is of great importance in clinical studies. Most of the existing approaches were based on traditional computer vision algorithms or conventional machine learning methods. Recently, deep learning also took its place on tackling this issue \cite{Chen:16}.

We developed a method of lymphocyte segmentation, which was based on the proposed XAE+FCN architecture. The process steps included: 1) lymphocyte detection; 2) lymphocyte classification; and 3) the integration with general image segmentation procedure provided in CellProfiler \cite{Carpenter:06:R100}. In this experiment, a lymphocyte H\&E image dataset \footnote{\url{http://www.andrewjanowczyk.com/deep-learning/}} was obtained from the online supplemental materials of the article \cite{Janowczyk:16:29}. The details of these steps will be described in the following sub-sections.

\subsubsection{Lymphocyte Detection}

In this step, first, we cropped lymphocyte images according to the annotations provided by the dataset. Since in the dataset, the negative (image background) samples were not provided, we used the same approach proposed by Janowczyk \textit {et al.} \cite{Janowczyk:16:29}, that is, cropping image patches with the exclusion locations annotated as lymphocyte based on a Baysan's na\"{i}ve classifier. Then, we performed XAE+FCN for the lymphocyte / background classification. Some image patterns can be found in Fig.~\ref{fig:lymphocyte_detection_backgrd} and \ref{fig:lymphocyte_detection_lympho}. The corresponding XAE features are showing in Fig.~\ref{fig:lymphocyte_detection_weights}.

The ROC plot and the confusion matrix can be found in Fig.~\ref{fig:lymphocyte_detection_roc} and Table~\ref{tbl:lymphocyte_detection_conf}, respectively. The performance was compared with the cutting-edge technology reported in \cite{Janowczyk:16:29}, in which, Janowczyk \textit{et al.} mentioned the F-score of $90\%$ of their approach, while the F-score of our solution reached $98.67\%$ (based on 10-cross-validation). Chen \textit{et al.} proposed another method, where the accuracy information of F-score and the complete dataset were not provided \cite{Chen:16}.

Although the accuracy of the proposed lymphocyte detection was excellent, however, a major drawback was, most of the false-positives were other cell types (see Fig.~\ref{fig:lymphocyte_classification_false}). This fact defeated the purpose of lymphocyte segmentation. Thus, we improved the results by performing an additional lymphocyte classification.

\subsubsection{Lymphocyte Classification}

In this step, we performed false-lymphocyte / true-lymphocyte classification. The training patterns for trye-lymphocytes were the same as in the previous step. The false-lymphocytes were the false-positives from the results of the previous step (see Fig.~\ref{fig:lymphocyte_classification_false} and \ref{fig:lymphocyte_classification_lympho}). Here we performed another XAE+FCN. The corresponding XAE features can be found in Fig.~\ref{fig:lymphocyte_classification_weights} and the ROC plot and the confusion matrix are showing in Fig.~\ref{fig:lymphocyte_classification_roc} and Table~\ref{tbl:lymphocyte_classification_conf}, respectively. The obtained F-score was 92.39\%.

\subsubsection{Integration}

Finally, we integrated these steps into a CellProfiler \cite{Carpenter:06:R100}, with its function blocks of general purpose bioimage segmentation. The scenario can be found in Fig.~\ref{fig:lympho_segmentation}. 

We used the lymphocyte contour data provided in the supplemental materials of  \cite{Janowczyk:16:29}, which includes the manually annotated contours of a subset of all labeled lymphocytes. 
The obtained Dice coefficient \cite{Dice:45:297} was $88.31\%$, while the best reported performance based on the same dataset was $74\%$ \cite{Kuse:10:235}.

\section{Conclusions}

\label{sec:conclusions}

In this paper, a novel architecture of autoencoder (AE), named exclusive autoencoder (XAE), was discussed for performing nucleus detection and classification.
In our experiments of biomedical image analysis, we studied the XAE applications on nucleus detection on breast cancer hematoxylin and eosin (H\&E) histopathological images, as well as multi-class nucleus classification, including epithelial, inflammatory, fibroblast and miscellaneous nucleus images on H\&E images. For nucleus detection, simply replacing AEs by XAEs, the performance index of F-score was significantly improved to from $84.49\%$ to $96.64\%$ on the same dataset. For nucleus classification, the obtained F-score was $70.4\%$. In the proposed lymphocyte segmentation, we have compared with cutting-edge technology, and gained the improved from $90\%$ to $98.67\%$ on the same dataset.

We also proposed an approach of lymphocyte segmentation which integrated the XAE+FCN algorithm with the general purpose bioimage segmentation method provided in CellProfiler and pushed the performance index of Dice coefficient from $74\%$ to $88.31\%$ on the same dataset. These results suggested the performance of existing applications of classification based on AEs, can further be improved by replacing AEs with XAEs, as long as the data labels are also available during the phase of feature extraction. 

In the experiment of nucleus classification, the proposed XAE+FCN approach didn't show the performance as good as in other experiments. One of the major reason was the morphological variation. More precisely, training patterns were not aligned. This also the conceptual idea of the work proposed by Sirinukunwattana \textit{et al.} \cite{Sirinukunwattana:16:1196}, where the locality sensitive features have been considered. It will be an interesting idea of integrating the proposed XAE+FCN into a locality sensitive neural network architecture.


{\small
\bibliographystyle{ieee}
\bibliography{main}
}

\end{document}